\def\year{2021}\relax
\documentclass[letterpaper]{article} 
\usepackage{aaai21}  
\usepackage{times}  
\usepackage{helvet} 
\usepackage{courier}  
\usepackage[hyphens]{url}  
\usepackage{graphicx} 
\usepackage{amsmath}
\usepackage{amssymb}
\usepackage{gensymb}
\usepackage{hyperref}
\usepackage{natbib}  
\usepackage{caption} 
\title{Playing Angry Birds with a Domain-Independent PDDL+ Planner}
\author{
    Wiktor Piotrowski,\textsuperscript{\rm 1}
    Roni Stern,\textsuperscript{\rm 1,2}
    Matthew Klenk,\textsuperscript{\rm 1}
    Alexandre Perez,\textsuperscript{\rm 1}
    Shiwali Mohan\textsuperscript{\rm 1} 
    Johan de Kleer,\textsuperscript{\rm 1}
    Jacob Le,,\textsuperscript{\rm 1} 
    \\
}
\affiliations{
    \textsuperscript{\rm 1} Palo Alto Research Center, CA, USA\\
    \textsuperscript{\rm 2} Ben-Gurion University of the Negev, Beer-Sheva, Israel\\
    e-mail: \{wiktorpi,rstern,aperez,smohan,dekleer,jale\}@parc.com;klenk.matt@gmail.com\\
}
\begin{document}

\maketitle

\begin{abstract}
This demo paper presents the first system for playing  the popular Angry Birds game using a domain-independent planner. 
Our system models Angry Birds levels using PDDL+, a planning language for mixed discrete/continuous domains. It uses a domain-independent PDDL+ planner to generate plans and executes them. 
In this demo paper, we present the system's PDDL+ model for this domain, identify key design decisions that reduce the problem complexity, and compare the performance of our system to model-specific methods for this domain. The results show that our system's performance is on par with other domain-specific systems for Angry Birds, suggesting the applicability of domain-independent planning to this benchmark AI challenge. 
A video explaining and demonstrating the behavior of our system is available at \texttt{https://bit.ly/35065UZ}. 
\end{abstract}



\section{Background: Angry Birds}
Angry Birds is a wildly popular mobile game in which the objective is to destroy pigs by launching birds at them from a slingshot. The pigs may be protected by structures built from blocks of various types. Each level of the game has an allotted number of birds that can be used, and pigs that need to be killed in order to solve the level.
Players earn points by destroying pigs and blocks as well as for each unused bird after the level has been solved.
This game is an interesting testbed for automated planning research since it 
requires reasoning about sequential actions with complex dynamics~\cite{renz2019ai}. 
Unlike classical planning, the agent's actions often triggers a cascading chain of events causing drastic changes to the world. 
The Angry Birds game is inherently a non-trivial hybrid system, exhibiting both discrete and continuous behavior, which includes non-linear dynamics. 
Game levels often contain a large number of objects, which vary in their type and functionality.
From a computational complexity perspective, different versions of the game have proven to be NP-Hard, PSPACE-complete, and EXPTIME-hard~\cite{stephenson2020computational},
Most existing AI agents for this game are based on encoding domain-specific strategies and selecting from them~\cite{borovicka2014datalab,wang2017description}. 
We describe the first successful application of domain-independent planning to solve Angry Birds, 
and demonstrate that its performance is on par with existing domain-specific approaches.

\section{Solving Angry Birds with a PDDL+ Planner}


Our agent, called Hydra, models Angry Birds levels using PDDL+ \cite{fox2006modelling}. PDDL+ is a planning language that enables modeling problems exhibiting both discrete mode switches and continuous flows. 
Hydra accepts an AngryBirds level from the Science Birds API~\cite{renz2019ai}, in the form of a list of labeled objects and their locations, and translates it to a PDDL+ problem. 
Then, it solves this problem using  UPMurphi~\cite{della2009upmurphi}, a well-known domain-independent PDDL+ planner. 

The objects in our PDDL+ model of Angry Birds are separated into four types: \textit{birds, pigs, blocks, and platforms}. 
Objects' properties such as locations, width, height, and velocities are modeled as functions. 
The slingshot, which is where the birds are launched from, is not modeled explicitly as an object in the domain. Instead, we assigned the slingshot's coordinates to every bird before it is launched.
Avoiding modeling unnecessary objects  
simplifies planning. 
The PDDL+ domain of Angry Birds features a variety of dynamics which dictate the change and evolution of the system. 
We modeled these dynamics through PDDL+ \emph{actions, events, and processes} constructs.  
PDDL+ events represent the system's mode switches which instantaneously change the dynamics of the modeled system, whereas PDDL+ processes represent processes that evolve the system over time, dictated by a set of ordinary differential equations.


\noindent\textbf{Actions} 
Angry Birds features two types of actions: (1) launching a bird from the slingshot at a chosen angle, and (2) activating a bird's special power (if available). 
Currently, we only consider the basic behavior of birds without their special powers. 
Launching of the bird consists of 
choosing an angle, adjusting the launch velocity, and releasing the bird. 
To mitigate the complexity of the launch action we fix the launch velocity to its maximum possible value, removing the need for the agent deliberating over the power adjustment. 
This decision is motivated by the fact that maximum velocity shots are the most common as they provide widest range of targets and impact velocity is proportional to damage. 
In addition, we paired the release action with a supporting process that continuously adjusts the angle as soon as a new bird is placed on the slingshot and is ready to be launched. 
Thus, to launch a bird the agent only needs to choose \textit{when to release it} instead of \textit{at which angle to do so}.

\noindent\textbf{Events} 
The vast majority of the Angry Birds system mechanics are event-driven. 
Each object in the game has one or more associated events which govern their interactions with the environment and other objects in the level.
They enable modeling of complex behavior including all collisions, destructions, explosions, and structure collapses. Events also model supporting features such as loading the next bird into the slingshot or exploding the currently active bird. 
Modeling the motion of blocks after impact and chains of block-block collisions, requires defining a process to track each individual block's change in position and rotation over time, as well as a set of events to account for secondary interactions. 
Keeping track of dozens of such non-linear processes in every state after a collision would significantly slow down the planner, while the improvement in the domain's accuracy is expected to be small. 
Therefore, we currently avoid modelling falling blocks processes and block-block collision events. 

\noindent\textbf{Processes}
There are two processes in our PDDL+ model: \textit{increase\_angle} and \textit{flying}.  The \textit{increase\_angle} process is a linear increase of the active bird release angle prior to launching it, as described above. This process is triggered as soon as the next bird in the queue is available for launch. 
The \textit{flying} process models the flight of the active bird, updating its velocity and location over time, according to the governing non-linear equations of motion. This process is triggered after a bird is released from the sling. 

\noindent\textbf{Simplified Modeling} While the goal of Angry Birds is to destroy all of the pigs, finding a solution to the corresponding PDDL+ problem was too difficult for our PDDL+ planner in some cases. 
Therefore, Hydra also generates two simplified PDDL+ problems:
(1) a \textit{single-shot problem}, which splits the original scenario into single-bird episodes with the goal of killing at least one pig, and (2) a \textit{single-shot-no-blocks problem}, which only considers platforms and pigs. In our current implementation, we first attempt to generate plans for the \textit{single-shot} problem. 
If no plan is found in 30s, we halt the planner and attempt to generate plans for the \textit{single-shot-no-blocks} problem. 
If again, no plan is found in 30s, we execute a pre-defined default action.

\vspace{-0.2cm}
\section{Experiments}

\begin{table}[tb!]
\centering
\small
\begin{tabular}{|c|c|c|}
\hline
\textbf{Agent} & \textbf{Problems Solved} & \textbf{Avg. Score per Level} \\ \hline
ANU    & \textbf{49}              & 45,112                      \\ \hline
Hydra          & 44                       & \textbf{47,516}             \\ \hline
DataLab        & 26                       & 46,668                      \\ \hline
Eagle's  Wing  & 16                       & 39,152                      \\ \hline
\end{tabular}
\caption{Levels solved and avg. scores given a 60 sec. time limit.}
\vspace{-0.5cm}
\label{tab:results}
\end{table}


  
We evaluated Hydra against two state-of-the-art agents: 
DataLab~\cite{borovicka2014datalab}, and
Eagle's Wing~\cite{wang2017description}. 
DataLab won the 2014 \& 2015 ScienceBirds competition and Eagle's Wing won the 2016-2018 competitions. 
Both agents work by trying a set of pre-defined strategies (e.g., destroy pigs, target TNT, target round blocks, destroy structures) and choosing the strategy that will yield the maximal predicted utility. 
Eagle's Wing also uses a pre-trained XGBoost model to optimize its performance. 
All agents are designed to work with the ScienceBirds API. For the evaluation, each agent attempted to solve a set of 100 randomly generated Angry Birds levels, which can be found at~\emph{gitlab.com/aibirds/sciencebirdsframework}. 
We compared the results of all agents to a baseline performance published by ANU, which was computed by averaging the scores of 
DataLab, 
Eagle's Wing, 
and an ANU-created naive agent. 
The latter targets a randomly selected pig with each active bird, and has predefined windows for apply special actions per bird type. It only considers the locations of visible pigs, disregarding all other objects in the scene.


Table~\ref{tab:results} shows the number of problem solved (out of 100) and the average score per level for each agent and the published baseline results (labelled ``ANU''). 
As can be seen, Hydra's performance is on par with existing domain-dependent approaches, even without encoding the birds' special powers. It solved more problems than DataLab and Eagle's Wing (44 vs. 26 and 16) and obtained the highest average score (47,516) per level. 
It is worth noting that both Eagle's Wing and DataLab's were designed for the ScienceBirds competition, where the levels were hand-crafted with specific winning strategies. In contrast, the levels we used were created by ANU's automated level generator.

\section{Conclusion}



We presented an agent that successfully plays the Angry Birds game  using a domain-independent planner.
Our agent translates Angry Birds levels to PDDL+ problems and solves them using a PDDL+ solver. 
This demonstrates the expressive power of PDDL+ and corresponding solvers. 
We plan on further extending our domain to improve its accuracy and develop domain-independent heuristics for more efficient PDDL+ planning. 
In addition, we are exploring techniques to diagnose incorrect PDDL+ models and repair them automatically~\cite{klenk2020model}. 




\small
\bibliography{main}
\end{document}